\documentclass[letterpaper]{article} 
\usepackage{aaai24}  
\usepackage{times}  
\usepackage{helvet}  
\usepackage{courier}  
\usepackage[hyphens]{url}  
\usepackage{graphicx} 
\usepackage{natbib}  
\usepackage{caption} 
\usepackage{amssymb}
\usepackage{array}
\usepackage{algorithm}
\usepackage{algorithmic}
\usepackage[utf8]{inputenc}
\usepackage{url}
\usepackage{booktabs}

\usepackage{bbding}
\usepackage{pifont}
\usepackage{wasysym}
\usepackage{utfsym} 
\usepackage{fontawesome}
\usepackage{multirow}

\makeatletter
\@namedef{ver@amsmath.sty}{}
\makeatother
\usepackage{amstext}

\usepackage{enumitem}

\usepackage{newfloat}
\usepackage{listings}
\usepackage[hyphens]{url}
\usepackage{url}
\usepackage{xcolor}
\definecolor{my_pink}{RGB}{210, 89, 151}

\DeclareCaptionStyle{ruled}{labelfont=normalfont,labelsep=colon,strut=off} 
\floatstyle{ruled}
\newfloat{listing}{tb}{lst}{}
\floatname{listing}{Listing}
\nocopyright 

\title{Boosting Multi-view Stereo with Late Cost Aggregation}
\author {
    Jiang Wu, 
    Rui Li,
    Yu Zhu\thanks{Corresponding author.},
    Wenxun Zhao,
    Jinqiu Sun,
    Yanning Zhang
}
\affiliations {
    Northwestern Polytechnical University\\
    {
    \textcolor{my_pink}{\url{https://github.com/Wuuu3511/LAMVSNET}}}
}

\begin{document}

\maketitle

\begin{abstract}
Pairwise matching cost aggregation is a crucial step for modern learning-based Multi-view Stereo (MVS). 
Prior works adopt an early aggregation scheme, which adds up pairwise costs into an intermediate cost. However, we analyze that this process can degrade informative pairwise matchings, thereby blocking the depth network from fully utilizing the original geometric matching cues.
To address this challenge, we present a late aggregation approach that allows for aggregating pairwise costs throughout the network feed-forward process, achieving accurate estimations with only minor changes of the plain CasMVSNet.
Instead of building an intermediate cost by weighted sum, late aggregation preserves all pairwise costs along a distinct view channel. This enables the succeeding depth network to fully utilize the crucial geometric cues without loss of cost fidelity. Grounded in the new aggregation scheme, we propose further techniques addressing view order dependence inside the preserved cost, handling flexible testing views, and improving the depth filtering process. Despite its technical simplicity, our method improves significantly upon the baseline cascade-based approach, achieving comparable results with state-of-the-art methods with favorable computation overhead.
\end{abstract}

\section{Introduction}
Multi-view stereo (MVS) has made great strides in image-based 3D reconstruction. At its core, the pairwise matching cost \cite{furukawa2015multi} represents essential depth cues
by matching over predefined depth hypotheses. 
With the advent of deep learning, the matching cost acts as the fundamental input of CNN-based methods \cite{yao2018mvsnet,gu2020cascade,wang2022mvster,li2023learning}, easing the depth network from geometric ambiguities and furthering the final estimation accuracy.
\par
\begin{figure}[t]
\begin{center}
\includegraphics[scale=0.3]{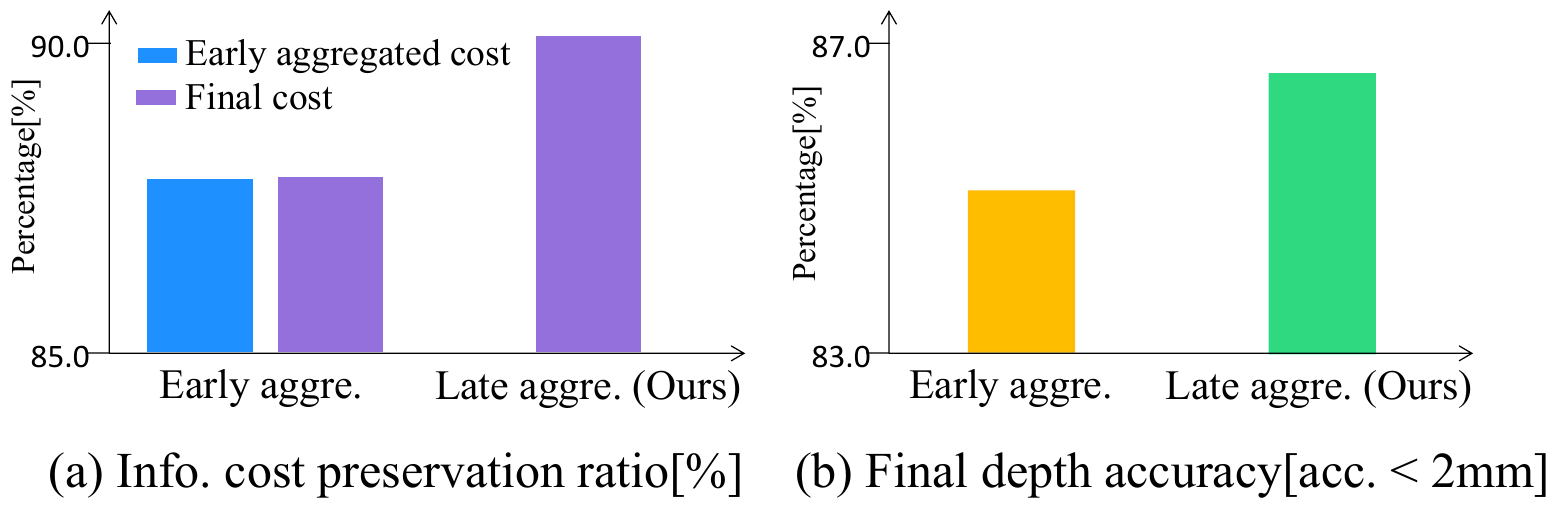}
\end{center}
\caption{\textbf{Comparison between early and late aggregation.} (a) We compute the reservation ratio, \textit{ie}, the proportion of pixels whose latent faithful depths are preserved by the final depth predictions. For the early aggregation method \cite{xu2022learning} (Early aggre.), its preservation ratio declines after the aggregation (blue bar), resulting in inferior network predictions (purple bar). While our method (late aggre.) effectively attains more informative matching costs via late aggregation. (b) As a result, our method achieves higher depth accuracy (acc.$<$2mm, the percentage of pixels within $2mm$ depth error) compared to the early aggregation method.
}
\label{fig:Pre_ratio}
\end{figure}
Though pairwise matching cost is crucial for CNN-based MVS, the pairwise matching costs from different view pairs are not equally informative, due to varying matching conditions such as illumination, occlusion as well as scene appearance. 
For a pixel with multiple pairwise matching costs, the \textit{informative} matching cost typically reflects the true scene depth cues, where the highest matching value is distinguished from the remaining values, and its corresponding depth hypothesis approximates the ground truth depth well. 
Meanwhile, there also exist \textit{non-informative} costs, where the matching cost shows non-distinct peaks or the corresponding depth hypothesis does not correlate to the ground truth depth.
Among the two cost types, the informative cost is vital for the depth network, due to its direct and faithful depth cues from well-conditioned geometric matching. 
\par
\begin{figure*}[t]
\begin{center}
\includegraphics[width=0.95\linewidth]{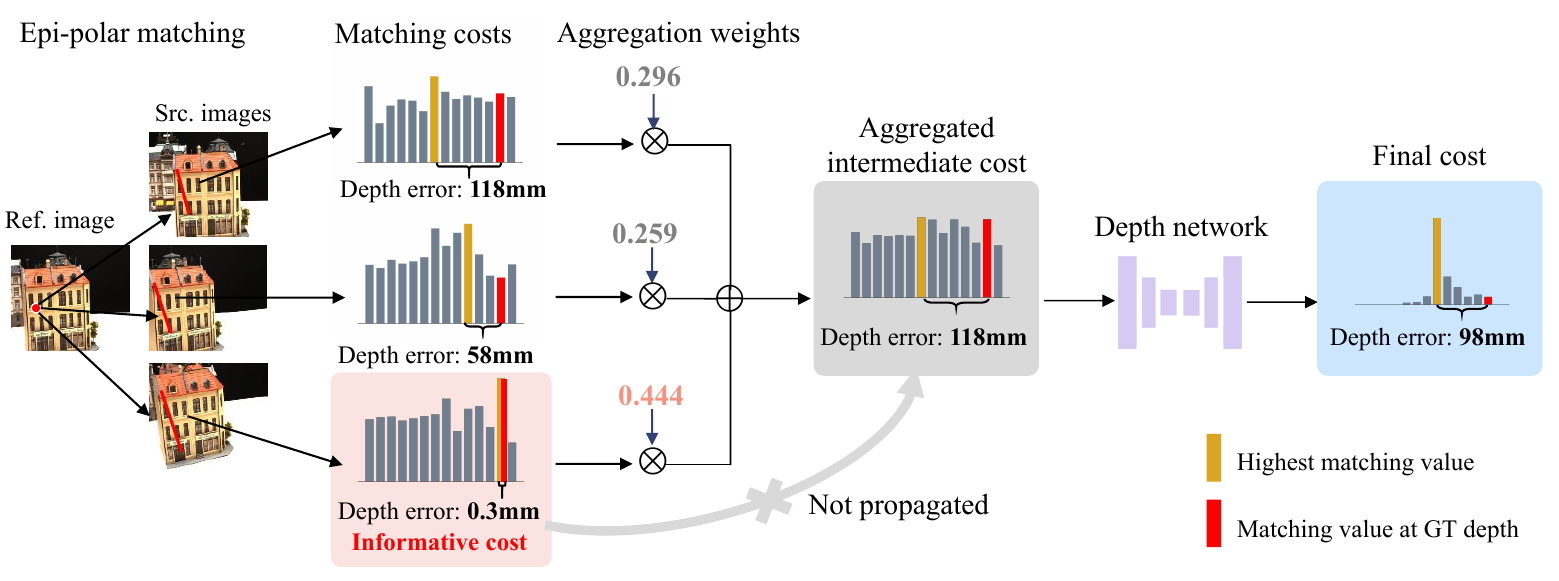}
\end{center}
\caption{\textbf{Limitations of early aggregation.} We showcase the early aggregation process of a pixel in PVSNet \cite{xu2022learning}. There are three pairwise matching costs with one informative cost having the most faithful depth cues (\textcolor{pink}{\textbf{pink}} background). Though the weight module managed to assign the informative cost with the highest weight, it still leads to a suboptimal intermediate cost (\textcolor{gray}{\textbf{gray}} background),  which compromises the informative cost with non-informative ones, due to indistinguishable weights. As a result, the depth network can not utilize the initial informative cost, leading to inferior depth predictions (\textcolor{blue}{\textbf{blue}} background).}
\label{fig:Lim_early_agg}
\end{figure*}

To make the best of the informative costs, previous methods \cite{zhang2023vis,xu2022learning,wang2022mvster}
typically adopt an early aggregation scheme, that multiple costs are merged into one intermediate cost early before the depth network. 
To let depth network access only the most informative ones, a standalone weight module \cite{zhang2023vis,xu2022learning,wang2022mvster} evaluates the faithfulness of each pairwise cost, followed by a weighted sum operation to aggregate costs. 
While these methods achieve good gains in challenging scenes, our analysis reveals a notable challenge: early aggregation can deteriorate the informative costs during the intermediate cost construction. 
Specifically, during the early aggregation, the vital depth cues from the informative matching cost can be compromised with the non-informative ones when the learned aggregation weights are not distinguishable enough (see Fig.\ref{fig:Lim_early_agg} and Section. 3).
Limited by the early aggregated cost, the vital depth cues from the informative cost may not propagate to the depth network, resulting in the under-utilization of geometric matching (see Fig.\ref{fig:Pre_ratio}).

\par
To address the revealed challenge, 
we advocate a late aggregation approach, which allows propagating and aggregating informative costs throughout the network forward pass. 
To fully utilize the initial informative costs, we preserve all pairwise costs along a view channel and send the view-preserved cost directly to the depth network.
This makes the aggregation ``late'' since the aggregation operates throughout the whole network forward pass instead of before the network. As such, the depth network benefits consistently from the initial informative matching costs that deliver faithful depth cues.
Grounded in the late aggregation, we further propose a view-shuffle technique to address the view order dependence within the preserved cost. We also propose two cost construction methods to handle flexible testing views for late aggregation, as well as an updated point cloud filtering strategy.
Despite the simplicity of our method, it suggests that better utilization of informative costs can be achieved without engineering weight modules and with only minor adjustments to the generic cascade-based MVS.
We conduct extensive experiments on the DTU, Tanks and Temples, as well as the ETH3D datasets. Our method significantly outperforms the baseline Cascade-based MVS, achieving comparable results with state-of-the-art methods with favorable computation overhead.
We hope our method can encourage further research on the aggregation design for multi-view stereo methods. 
The contributions of the paper are summarized as follows:
\par
\begin{itemize} 
\item We analyze the wide-adopted early aggregation scheme, and reveal that it can deteriorate the informative matching costs during intermediate cost construction, thus hindering the depth network from leveraging the faithful depth cues from geometric matching.
\item
In light of our analysis, we advocate the late aggregation that allows propagating and aggregating informative costs throughout the network forward pass. Grounded in the late aggregation, we further devise a set of techniques to alleviate reliance on view order, handle flexible test views, and achieve better point cloud filtering.
\item
We evaluate our method on DTU, Tanks and Temple, as well as the ETH3D datasets.
We achieved comparable performance to state-of-the-art methods without using their weight modules or updated depth networks.
\end{itemize}
\section{Related Work}
\subsection{Learning-based MVS}
In recent years, learning-based models have shown great advantages in multi-view reconstruction.
MVSNet~\cite{yao2018mvsnet} introduced a differentiable homography warping to build the cost volume, using 3D multi-scale U-Net for regularization.
However, due to its 3D CNN architecture, the memory and computation costs are quite expensive.
To allow handling high-resolution images, R-MVSNet~\cite{yao2019recurrent}, AA-RMVSNet~\cite{Wei_2021_ICCV} and DH2C-RMVSNet~\cite{yan2020dense} integrate the RNN-based recurrent models to MVS reconstruction.
They utilized GRU/LSTM to replace the 3D convolution.
Although high-resolution images can be processed, reasoning takes a lot of time.
Cascade-structured methods~\cite{cheng2020deep,gu2020cascade,yang2020cost,mi2022generalized} further improve the efficiency of the pipeline.
They predict a coarse depth map in the early stages and then gradually reduce the depth range and increase the image resolution for finer depth prediction. 
Furthermore, some methods aim to enhance the feature aggregation~\cite{ding2022transmvsnet,cao2022mvsformer} and geometric perception~\cite{zhang2023geomvsnet} by designing additional networks, further improving the performance of the network.
While these methods have achieved promising results, the current MVS pipeline still has some drawbacks in aggregating matching information of different views.
\subsection{Pairwise Costs Aggregation}
As the pairwise costs are not equally informative due to varying matching conditions, aggregating the pairwise cost to make the best of the matched depth cues is crucial for MVS methods. Early deep learning-based multi-view stereo (MVS) methods~\cite{yao2018mvsnet, gu2020cascade} use variance for early aggregation of pairwise cost volumes which treat each viewpoint equally, ignoring visibility information~\cite{xu2022learning} and limiting the performance of the depth network. 
Currently, visibility~\cite{xu2022learning,zhang2023vis,wang2022mvster} weight-based early aggregation has become mainstream in MVS methods.
PVSNet~\cite{xu2022learning} utilizes pairwise costs to regress the visibility map as weighting guidance for the summation of the pairwise volumes.
Vis-MVSNet~\cite{zhang2023vis} explicitly infers occlusion information from matching uncertainty estimation and inferred the pairwise uncertainty map jointly with the pairwise depth map.
MVSTER~\cite{wang2022mvster} introduces the epi-polar Transformer to compute visibility maps without bringing additional parameters.
Besides early aggregation with CNN-based weight modules, the recent method SimRecon~\cite{sayed2022simplerecon} adopts an MLP-based aggregator to combine the pairwise costs into an intermediate cost before the depth network.
Although the early aggregation used by the above methods is effective in challenging scenes, as revealed by our analysis in the next section, the informative matching costs among all pairwise match costs can deteriorate during the intermediate cost construction. 
In this paper, we advocate a late aggregation approach that encourages the informative matching costs to propagate and aggregate throughout the network forward. 
Our approach distinguishes from previous methods in the obviation of intermediate cost reconstruction, allowing the depth network to benefit from the initially matched depth cues throughout the network forward process.
\section{Limitation of Early Aggregation}\label{ama}
As the informative matching costs are essential for learning-based MVS, in this section, we evaluate the early aggregation by measuring its capability of preserving informative costs. 
Then we analyze the early aggregation's limitation, which inspires the further design of late aggregation.
\par
\noindent \textbf{Preservation ratio of informative costs.} To evaluate the capability of preserving informative matching costs, we access if pixels with the informative matching cost can yield accurate final depth after the network forward pass. Herein, a pixel is regarded to have an informative matching cost if one of its pairwise costs' highest matching value corresponds to the ground truth depth hypothesis. With this criterion, 1) we first extract pixels with informative matching costs for each target image, then 2) we compute the ratio of the pixel subset whose final depth prediction still lies in the ground truth depth hypothesis. 
This ratio denotes the portion of initial informative matching costs being utilized by the depth network. Results on the DTU validation set are shown in Fig. \ref{fig:Pre_ratio}, where we provide the results of PVSNet \cite{xu2022learning} with early aggregation (Early aggre.). We find from Fig. \ref{fig:Pre_ratio}.(a) that the preservation ratio of PVSNet started declining within the early aggregated cost (the \textcolor{blue}{\textbf{blue}} bar), and the resulting final estimation does not yield higher preservation ratio despite using parameter-intensive depth network (the \textcolor[RGB]{147,112,219}{\textbf{purple}} bar). It appears that the early aggregated cost with declined informative cost proportion sets up an upper bound for the final estimation, which hinders the depth network from leveraging all informative matching costs.
\noindent \textbf{Analysis of the information loss for early aggregation.} Observing the decline of informative costs during early aggregation, we further analyze the reason behind this. As shown in Fig. \ref{fig:Lim_early_agg}, we showcase PVSNet's \cite{xu2022learning} early aggregation over three pairwise matching costs of a pixel. The initial matching costs contain one informative matching (highlighted in \textcolor{pink}{\textbf{pink}}) whose highest matching value corresponds to the ground truth depth. The weight module takes three matching costs and assigns weights gauging the faithfulness of each. As seen in the figure, though it assigns the informative costs with the highest weight (0.444), the weights are not distinguishable enough over the rest (0.296, 0.259). Consequently, the aggregated intermediate (highlighted in \textcolor{gray}{\textbf{gray}}) cost delivers depth cues far from the informative matching cost due to the accumulated error. With only access to the aggregated cost, the depth network can not regress to the correct depth value in final cost (highlighted in \textcolor{blue}{\textbf{blue}}), even if the geometric matching does provide high-quality depth cues. 
The analysis motivates us to advocate a different aggregation scheme, that the depth network can further benefit from the informative costs if they can be preserved and aggregated throughout the forward process, which we call a late aggregation scheme. 
As shown in Fig. \ref{fig:Pre_ratio}, by adopting the late aggregation, our method achieves both higher preservation ratio (Fig. \ref{fig:Pre_ratio} (a)) and better final accuracy compared to the early aggregation scheme (Fig. \ref{fig:Pre_ratio} (b)).
\section{Methodology}
Given a reference image $\mathbf{I}_0 \in \mathbb{R}^{H\times W \times 3}$ and $N-1$ source images $\{\mathbf{I}_{i}\}{_{i=1}^{N-1}}$, as well as camera intrinsics parameters $\{\mathbf{K}_{i}\}_{i=1}^{N-1}$ and extrinsic parameters $\{[\mathbf{R}_{0\rightarrow i};\mathbf{t}_{0 \rightarrow i}]\}_{i=1}^{N-1}$, our goal is to predict the depth map of the reference image $\mathbf{I}_0$. 
In this section, we first introduce the late aggregation scheme, as well as the view shuffle approach to address view dependence for late aggregation. We then introduce two techniques to accommodate flexible testing inputs.
Finally, we introduce an updated filtering scheme to yield an accurate final point cloud.
\begin{figure}[t]
\begin{center}
\includegraphics[scale=0.44]{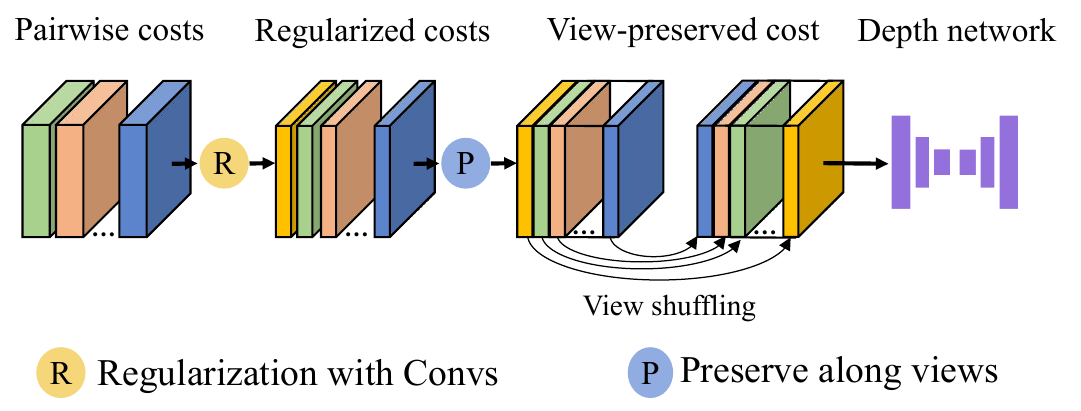}
\end{center}
   \caption{\textbf{Late cost aggregation with view-preserving.}
   For multiple pairwise costs, we apply a single convolutional layer separately to each cost for pre-regularization. Then, we preserve the pairwise costs along a view-based channel to build view-preserved costs for the depth network. To disentangle the view order dependence with late aggregation, we introduce the view shuffle scheme aimed at disrupting the view order.
   }
\label{fjg::vpa_vs}
\end{figure}
\subsection{Late Cost Aggregation}\label{svc}
\noindent{\textbf{Pairwise costs construction}}
Like previous MVS methods, we sample a set of depth hypotheses $d\in\{d_{j}\}_{j=1}^{M}$ within a pre-determined depth range \cite{yao2018mvsnet}. 
Using a differentiable homograph, we can align the features of source views to the reference view
\begin{equation}
\mathbf{p}'=\mathbf{K}_{0}[\mathbf{R}_{0\rightarrow i}(\mathbf{K}_{i}^{-1}\mathbf{p} d)+\mathbf{t}_{0\rightarrow i}],
\end{equation}
where $\mathbf{R}$ and $\mathbf{t}$ denote the rotation and translation parameters. $\mathbf{K_{0}}$ and $\mathbf{K_i}$ are the intrinsic matrices of the reference and source cameras.
Source features $\mathbf{F}{_i}$ at pixel $\mathbf{p}$ can be warped to the position of coordinate $\mathbf{p}'$ and interpolated to the original resolution.
The pairwise volumes $\{\mathbf{V}_{i}\}_{i=1}^{N-1}$ are then constructed based on the dot product similarity
\begin{equation}
\mathbf{V}_{i}^{d}(\mathbf{p}')=\mathbf{F}_{0}(\mathbf{p}') \odot \mathbf{F}_{i}^{d}(\mathbf{p}),
\end{equation}
where $\mathbf{F}_{i}^{d}(\mathbf{p})$ denotes the warped $i$-th source feature map at depth $d$ and $\odot$ refers to the Hadamard product \cite{horn1990hadamard}.

\noindent{\textbf{Late aggregation with preserved views.}}
As discussed before, early aggregation can deteriorate the informative matching costs during the intermediate cost construction.
To address this issue, herein we advocate late aggregation, which obviates constructing an intermediate cost and allows the depth network to consistently benefit from the informative costs.
As shown in Fig. \ref{fjg::vpa_vs}, given each pairwise cost volume $\mathbf{V}_i$ $\in$ $\mathbb{R}^{{H'\times}W'{\times}D{\times}C}$ , we regularize it with a shallow convolution layer $g_{\theta}$, yielding $g_{\theta}(\mathbf{V}_i)$ of shape ${H'\times}W'{\times}D{\times}C'$.
We set $C'=1$ as we find no obvious improvements with more view channels.
To enable the utilization of informative matching costs for the depth network, we preserve all regularized pairwise cost volumes along a view dimension
\begin{equation}
C_{\text{VP}} = [g_{\theta}(\mathbf{V}_1),..., g_{\theta}(\mathbf{V}_{N-1})] \in \mathbb{R}^{H' \times W' \times D \times (N-1)},
\end{equation}
where $C_{\text{VP}}$ denotes the view-preserved cost volume. 
Benefiting from the view-preserved cost, the depth network can directly access and aggregate the informative costs from pairwise matching.
\begin{figure}[t]
    \centering  
	 \includegraphics[scale=0.3]{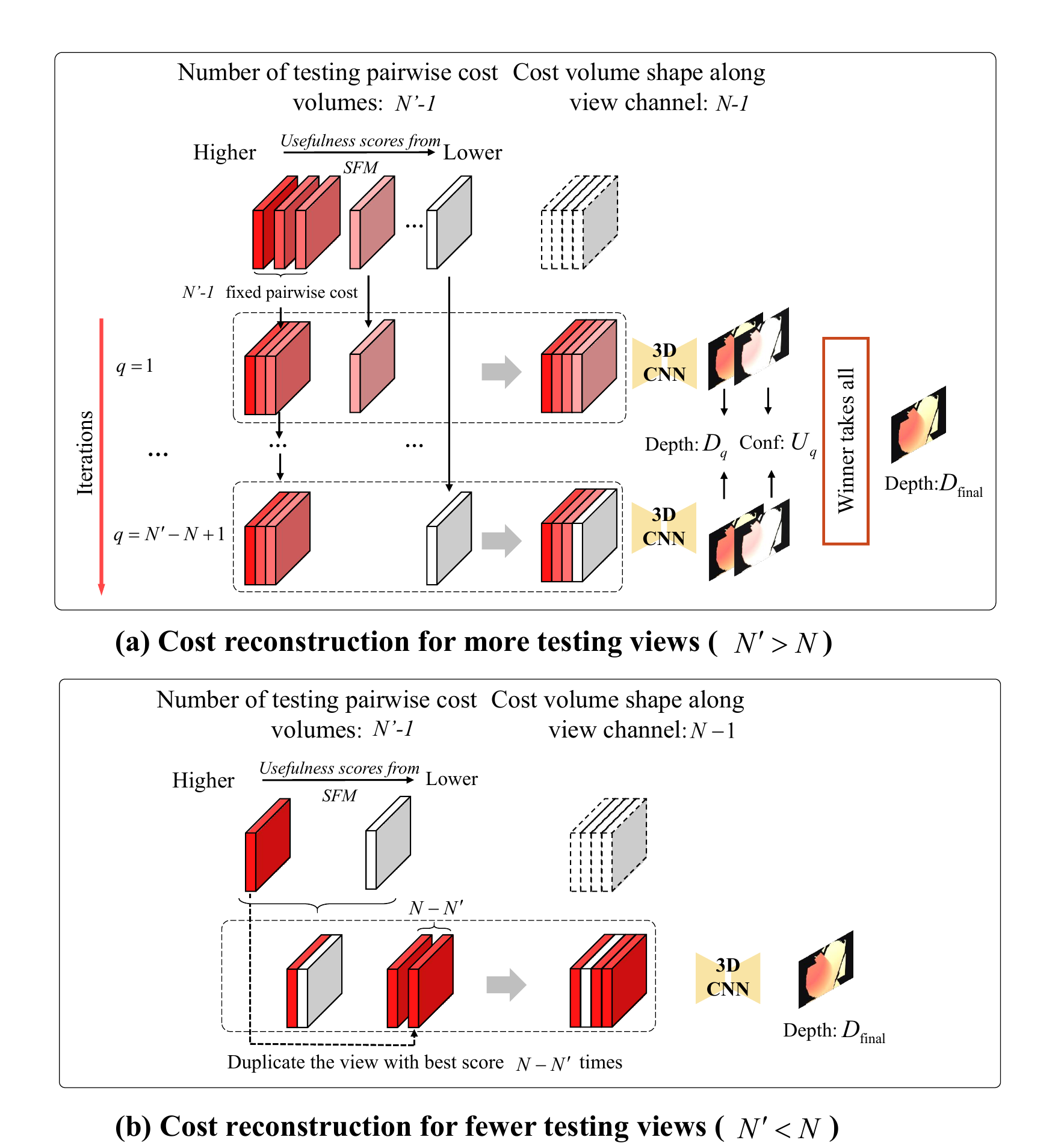}
	 \caption{\textbf{Cost reconstruction for handling flexible numbers of testing views.} (a) For more testing views, we select the most useful pairwise costs and iterate the remaining to reconstruct multiple cost that fits the initial cost shape. The final depth is yielded with the winner-takes-all strategy. (b) For fewer testing views, we keep all pairwise costs and duplicate the most useful one repetitively until the cost fits the desired shape.
  }
\label{fig:flextest}
\end{figure}
As the view-preserved cost contains matching cost from each view pair, it inevitably involves view order information. We found that the depth can be affected if the pairwise costs are preserved in a fixed order, \textit{e.g.}, from small to wide baselines. We disentangle this view-order dependence by incorporating randomness into the preserved cost, \textit{i.e.}, shuffling the view channels.
This steers the depth network to better leverage the informative costs across views.
\par
We note that simple as the view-preserving method is, it underscores that a technically simple approach supported by motivated analysis can lead to non-negligible improvement (as shown in Table. \ref{table::abl_in_dtu}). As such, better utilization of informative matching costs can be achieved without carefully engineered weight modules \cite{zhang2023vis,xu2022learning}, and with only minor adjustments on the plain cascade MVS.
\subsection{Handling Flexible Number of Testing Inputs}\label{BRR}
The view-preserved cost contains all pairwise costs, therefore its shape along the view channel depends on the training pairwise cost number $N-1$. In real-world applications, however, the testing pairwise cost number $N'-1$ varies, which may differ from $N-1$.
To address this issue, we propose two test-time cost construction methods that accommodate the varying number of testing-time pairwise costs.
\par
\noindent\textbf{Cost construction for more testing views ($N'>N$).}
To accommodate more testing pairwise costs into the fixed preserved cost shape, we propose an iterative inference scheme, where the pairwise cost volumes with potentially more informative pixel-wise costs are kept fixed, and the remaining pairwise cost volumes are selected iteratively.
Concretely, for each iteration $q$, we keep the top $N'-2$ pair-wise cost volumes respective to their usefulness scores given by the view selection strategy \cite{yao2018mvsnet}, and let the remaining $1$ view iterated through the remaining.
As shown in Fig. \ref{fig:flextest} (a), this leads to $N'-N+1$ iteration steps, each constructing a view-preserved cost with the same size as the training phase. Each iteration yields a depth map $D_{q}$ and a confidence map $U_{q}$ describing the confidence of each estimated depth.
We further compute the final depth $D_\text{final}$ by assigning each pixel the depth of the highest confidence score, among all $N'-N+1$ estimations.
\par
\noindent\textbf{Cost construction for fewer testing images ($N'<N$)}
When the testing pairwise costs are fewer than that of the training, we aim to duplicate the pairwise costs with potentially more informative pixel-wise costs to complement the view-preserved cost shape. As shown in Fig. \ref{fig:flextest} (b), we keep all pairwise costs from testing view pairs, as well as duplicating the pairwise cost with the highest usefulness scores $N-N'$ times to fit into the training-time cost shape. The final depth $D_\text{depth}$ can thus be regressed from the constructed cost.
With the proposed testing-time cost construction techniques, our method effectively handles arbitrary numbers of testing views (as shown in Tab. \ref{tab::viewnumbles}) in practical scenarios.
\subsection{Multi-view Consistency Filtering}\label{dmf}
Depth map filtering is important to improve the quality of generated point cloud. Traditional methods use photometric and geometric consistency constraints for filtering \cite{yao2018mvsnet}. However, previous filtering algorithms were unable to generalize to all scenes due to fixed constraints on the number of viewpoints.  ~\cite{yan2020dense} proposed dynamic consistency checking, which improved the completeness of the point cloud but introduced many noise points, especially for points with large absolute depth values. We improved upon the generation of masks in the dynamic consistency check algorithm by using absolute depth instead of relative depth and introducing photometric consistency constraints also from the source view, which enhances accuracy while ensuring the completeness of the point cloud.
\begin{figure}[t]
\begin{center}
\includegraphics[scale=0.33]{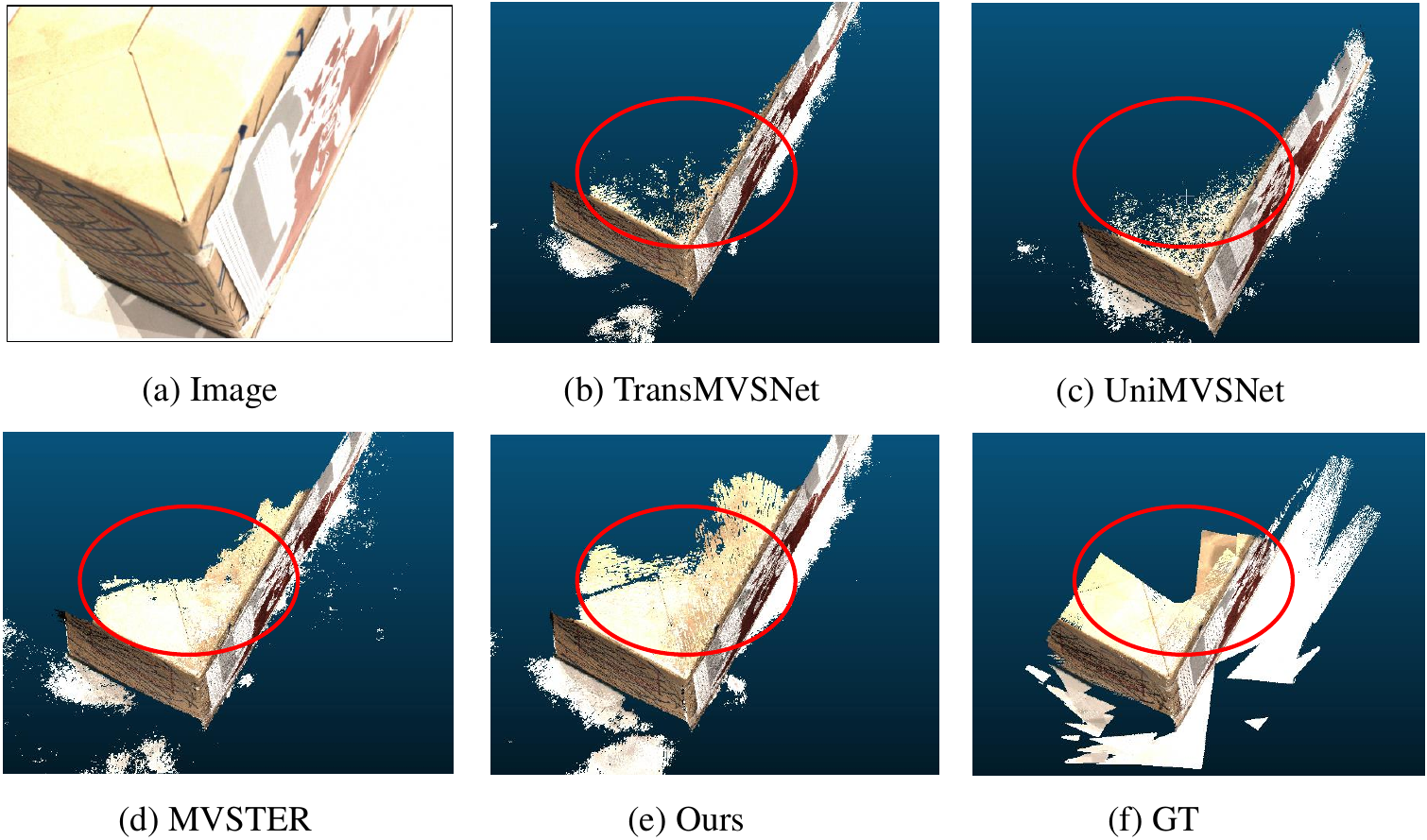}
\end{center}
   \caption{\textbf{Reconstruction of scan 13 on DTU}. 
   Our method achieves faithful reconstruction results, especially in textureless areas with high reflection. Note that (d) MVSTER uses an original resolution of 1600×1200 for network training, while other methods are trained on low-resolution images.}
\label{fig::dtu}
\end{figure}

\section{Experiments}
In this section, we evaluate our method and show its superiority on three major MVS benchmarks (DTU~\cite{aanaes2016large}, Tanks and Temple~\cite{knapitsch2017tanks}, ETH3D~\cite{schops2017multi}) over previous methods, and conduct extensive ablations to verify the effectiveness of the proposed contributions. 
\subsection{Datasets}
{DTU}~\cite{aanaes2016large} is an indoor data set under laboratory conditions with multi-view images, camera poses, and depth ground-truth, which contains 124 scenes with 49 views and 7 illumination conditions. 
Following MVSNet~\cite{yao2018mvsnet}, DTU is split into training, validation, and test set.
{BlendedMVS}~\cite{yao2020blendedmvs} is a large-scale dataset with indoor and outdoor scenes, containing 106 training scans and 7 validation scans. Like most MVS methods, we only use it for training.
To compare it with other state-of-the-art methods, we evaluate our model on its intermediate and advanced sets which consist of 8 and 6 scenes, respectively.
The {ETH3D~\cite{schops2017multi}} benchmark introduces high-resolution images characterized by challenging viewpoint variations, which are partitioned into distinct training and test sets.
\begin{table}[H]
 \centering
 \setlength{\tabcolsep}{2.2pt}
 \scalebox{0.75}{
\begin{tabular}{lccc}
\toprule
Method      & Accuracy & Comp. & Overall \\ \hline
Gipuma~\cite{galliani2015massively}      & \textbf{0.283}    & 0.873     & 0.578       \\
COLMAP~\cite{schonberger2016pixelwise}      & 0.400    & 0.664     & 0.532       \\
CasMVSNet~\cite{gu2020cascade}   & 0.325    & 0.385     & 0.355       \\
AA-RMVSNet~\cite{Wei_2021_ICCV}  & 0.376    & 0.339     & 0.357       \\
PVSNet~\cite{xu2022learning}      & 0.337    & 0.315     & 0.326       \\
Vis-MVSNet~\cite{zhang2023vis}  & 0.369    & 0.361     & 0.365       \\
MVSTER~\cite{wang2022mvster}      & 0.350    & 0.276     & 0.313       \\
UniMVSNet~\cite{peng2022rethinking}   & 0.352    & 0.278     & 0.315       \\
TransMVSNet~\cite{ding2022transmvsnet} & \underline{0.312}    & 0.298     & 0.305       \\
GbiNet~\cite{mi2022generalized}$*$   & 0.314    & 0.295     & 0.305             \\ 
GeoMVSNet~\cite{zhang2023geomvsnet}   & 0.331    & \underline{0.259}     & \textbf{0.295}             \\ 
\midrule
\textbf{Ours}        & 0.335    & \textbf{0.258}     & \underline{0.297}      \\ 
\bottomrule 
\end{tabular}}
\caption{Quantitative results on DTU~\cite{aanaes2016large} ({lower is better}). 
Our approach has demonstrated performance on par with established state-of-the-art method~\cite{zhang2023geomvsnet}.
The * indicates using unified thresholds as other methods for fair comparisons.}
\label{table::dtu}
\end{table}
\begin{table*}[]
 \centering
  \renewcommand{\arraystretch}{1.2}
 \scalebox{0.7}{
\begin{tabular}{lcccccccccccccccc}
\toprule
\multirow{2}{*}[-0.5ex]{Methods} & \multicolumn{9}{c}{Intermediate}                                     & \multicolumn{7}{c}{Advanced}                          \\ \cline{2-17} 
                         & Mean  & Fam.  & Fra.  & Hor.  & Lig.  & M60   & Pan.  & Pla.  & Tra.  & Mean  & Aud.  & Bal.  & Cou.  & Mus.  & Pal.  & Tem.  \\ 
\midrule
COLMAP~\cite{schonberger2016pixelwise}                   & 42.14 & 50.41 & 22.25 & 26.63 & 56.43 & 44.83 & 46.97 & 48.53 & 42.04 & 27.24 & 16.02 & 25.23 & 34.70 & 41.51 & 18.05 & 27.94 \\
CasMVSNet~\cite{gu2020cascade}                & 56.84 & 76.37 & 58.45 & 46.26 & 55.81 & 56.11 & 54.06 & 58.18 & 49.51 & 31.12 & 19.81 & 38.46 & 29.10 & 43.87 & 27.36 & 28.11 \\
Vis-MVSNet~\cite{zhang2023vis}               & 60.03 & 77.40 & 60.23 & 47.07 & 63.44 & 62.21 & 57.28 & 60.54 & 52.07 & 33.78 & 20.79 & 38.77 & 32.45 & 44.20 & 28.73 & 37.70 \\
MVSTER~\cite{wang2022mvster}               & - & - & - & - & - & - & - & - & - & 37.53 & 26.68 & 42.14 & 35.65 & 49.37 & 32.16 & 39.19 \\ 
EPP-MVSNet~\cite{ma2021epp}               & 61.68 & 77.86 & 60.54 & 52.96 & 62.33 & 61.69 & 60.34 & \textbf{62.44} & 55.30 & 35.72 & 21.28 & 39.74 & 35.34 & 49.21 & 30.00 & 38.75 \\
TransMVSNet~\cite{ding2022transmvsnet}              & 63.52 & 80.92 & 65.83 & 56.94 & 62.54 & 63.06 & 60.00 & 60.20 & \textbf{58.67} & 37.00 & 24.84 & 44.59 & 34.77 & 46.49 & 34.69 & 36.62 \\
GBiNet~\cite{mi2022generalized}                  & 61.42 & 79.77 & 67.69 & 51.81 & 61.25 & 60.37 & 55.87 & 60.67 & 53.89 & 37.32 & \textbf{29.77} & 42.12 & 36.30 & 47.69 & 31.11 & 36.93 \\
UniMVSNet~\cite{peng2022rethinking}                 & 64.36 & 81.20 & 66.43 & 53.11 & \textbf{64.36} & \textbf{66.09} & \textbf{64.84} & 62.23 & 57.53 & 38.96 & 28.33 & 44.36 & \textbf{39.74} & \textbf{52.89} & 33.80 & 34.63 \\  \midrule

\textbf{Ours}                      & \textbf{64.58} & \textbf{81.83} &\textbf{67.74} & \textbf{58.48} & 62.30 & 64.92 & 63.91 & 59.12 & 58.34 & \textbf{40.12} & 29.40 & \textbf{45.61} & 38.55 & 51.69 & \textbf{35.16} & \textbf{41.87} \\ \bottomrule \\
\end{tabular}}
\caption{\textbf{Quantitative results of F-score on Tanks and Temples benchmark.}
Our method achieves the best overall results on the "Intermediate" splits, as well as the challenging "Advanced" split which has larger view variations.}
\label{table:tnt}
\end{table*}
\subsection{Implementation Details}
We first train and evaluate our model on the DTU~\cite{aanaes2016large} dataset, and then finetune it on the BlendedMVS~\cite{yao2020blendedmvs} dataset for Tanks and Temple~\cite{knapitsch2017tanks} and ETH3D~\cite{schops2017multi} evaluation, following the previous practice.
We follow the source-view selection strategy and depth interval division strategy proposed by MVSNet~\cite{yao2018mvsnet} and use the dataset preprocessed by MVSNet for our experiments.
Our framework is based on the multi-stage cascaded structure proposed by CasMVSNet~\cite{gu2020cascade}, \textit{without} any network alternations \cite{ding2022transmvsnet} or additional loss functions \cite{peng2022rethinking}.
Our method consists of three stages, each of which generates outputs with resolutions of 1/4, 1/2, and full image resolution, respectively.
The depth sampling of each stage is 48, 32, 8 with corresponding depth intervals of 4, 1, and 0.5.
At the training phase, we set the input image size to 512 ${\times}$ 640 and the number of input images to 5.
We train our model for 16 epochs using the Adam optimizer. 
The initial learning rate was set to 0.001 and decayed by a factor of 2 after the 10th, 12th, and 14th epochs.
During the fine-tuning phase, we use the BlendMVS dataset preprocessed by PatchMatchNet~\cite{wang2021patchmatchnet}. 
The input image size is set to 768 ${\times}$ 576 and the input view number is 9.
When inferring large-scale scenes~\cite{schops2017multi, knapitsch2017tanks}, inspired by UniMVSNet~\cite{peng2022rethinking}, we doubled the number of depth bins to obtain a finer-grained depth sampling.
\subsection{Benchmark Performance}
\paragraph{Evaluation on DTU dataset}

\begin{figure*}
\begin{center}
\includegraphics[scale=0.55]{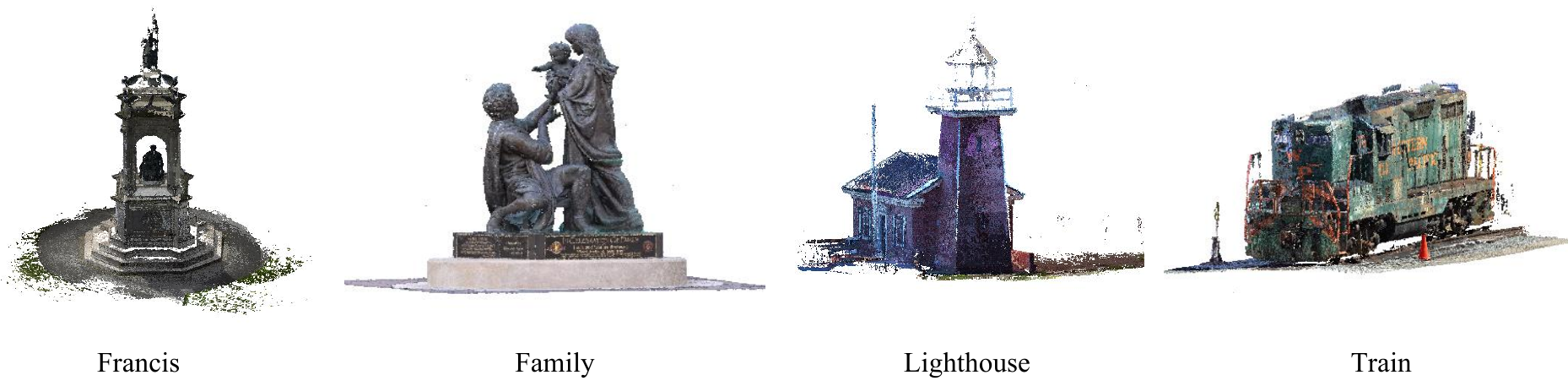}
\end{center}
\caption{\textbf{Qualitative results on Tanks and Temples.} Our method achieves detailed and complete reconstructions across different scenes.}

\label{fig::tnt}
\end{figure*}
We evaluated our proposed method on the test set of the DTU~\cite{aanaes2016large} dataset using official evaluation metrics, which compares the distance between ground-truth point clouds and the produced point clouds.
We set the number of input images to 5 with the resolution of 864 ${\times}$ 1152.
Quantitative results are shown in Tab. \ref{table::dtu}. With the simple technical updates on CasMVSNet \cite{gu2020cascade}, our method outperforms it with a significant improvement margin. Meanwhile, our method achieves comparable performance with the SOTA method GeoMVSNet~\cite{zhang2023geomvsnet}, with fewer parameters and higher speed (see Tab. \ref{table::time_parms}).
Quantitative results are shown in Fig. \ref{fig::dtu}, our generated point cloud has better completeness compared with previous methods, indicating a better utilization of the geometric matchings.

\begin{table}[]
\renewcommand{\arraystretch}{1.2}
\centering
\setlength{\tabcolsep}{2.2pt}
\scalebox{0.7}{
\begin{tabular}{lcccccc}
\toprule
\multirow{2}{*}[-0.5ex]{Methods} & \multicolumn{3}{c}{Training} & \multicolumn{3}{c}{Test}      \\ \cline{2-7} 
                         & Acc.          & Comp.           & F-score              & Acc.            & Comp.            & F-score \\ \hline
COLMAP~\cite{schonberger2016pixelwise} & \textbf{91.85}   &55.13            &67.66                 &\textbf{91.97}   &62.98             &73.01              \\ 
ACMM~\cite{xu2019multi}                  & 90.67          &70.42            &78.86                 &90.65            &74.34             &80.78              \\ \midrule
IterMVS~\cite{Wang_2022_CVPR}              & {73.62}   &61.87            &66.36                 &76.91   &72.65             &74.29              \\ 
EPP-MVSNet~\cite{ma2021epp}             & 82.76           &67.58            &74.00                 &85.47            &81.79             &83.40              \\ 
GBi-Net~\cite{mi2022generalized}                  & 73.17         &69.21            &70.78                 &80.02           &75.65             &78.40   \\ 
UniMVSNet~\cite{peng2022rethinking}               & {84.95} &60.62            &69.53                 &86.87   &77.58             &81.60              \\ 
TransMVSNet~\cite{ding2022transmvsnet}             & 79.05          &57.83            &64.74                 &84.02            &70.07             &75.39              \\ 
MVSTER~\cite{wang2022mvster}                  & 76.92          &68.08            &72.06                 &77.09            &82.47             &79.01   \\ 
PVSNet~\cite{xu2022learning}                  & 83.00          &71.76            &76.57                &81.55            &83.97             &82.62   \\ 
Vis-MVSNet~\cite{zhang2023vis}              & 83.32          &65.53   & {72.77}       &86.86            &80.92    &83.46              \\ \midrule
Ours  & 78.12        &\textbf{77.31}   & \textbf{77.27}                  & 83.94             &\textbf{84.38}          &\textbf{83.57}               \\ \bottomrule \\
\end{tabular}
}
\caption{\textbf{Quantitative results on ETH3D dataset}. Comparisons of reconstructed point clouds on ETH3D using percentage metric (\%) at threshold 2cm (higher is better). Our approach achieves the best performance on ETH3D datasets, which has challenging large viewpoint variations.}\vspace{-11pt}
\label{table::eth3d}
\end{table}
\paragraph{Evaluation on Tanks and Temples dataset.} We evaluate the generalization of our model on the Tank and Temple
datasets~\cite{knapitsch2017tanks}.
We use the original resolution as input and test our method with 11 viewpoints.
Results are shown in Tab. \ref{table:tnt}.
Our method achieves the best overall performances in both "Intermediate" and "Advanced" splits. It is worth noticing that the improvement is even higher in the more challenging "Advanced" split that has larger viewpoint variations. It shows the effectiveness of our method to generalize well to the challenging scenes with the proposed late aggregation scheme.

\paragraph{Evaluation on ETH3D dataset.} To further evaluate the robustness of the proposed method, we compare different methods on the challenging ETH3D dataset, where more significant viewpoint changes can produce a larger proportion of non-informative costs. The number of input views is 11 and the resolution of images is resized to 1080$\times$2048 or 2048$\times$3072. As shown in Tab. \ref{table::eth3d}, our method achieves the best performance in both the training and testing split. Note that our method outperforms the previous early-aggregation-based methods PVSNet \cite{xu2022learning} and Vis-MVSNet \cite{zhang2023vis} that adopt well-engineered weight modules to handle large viewpoint variations. The results show an encouraging clue that better utilization of informative costs under challenging scenarios can be achieved without well-engineered weight modules.
\begin{table}
 \centering
 \setlength{\tabcolsep}{6.5pt}
\begin{tabular}{cccccccc}
\toprule
\multicolumn{5}{c}{\textbf{Model Settings}} & \multicolumn{3}{c}{\textbf{Mean Distance}} \\ \midrule
 &  & LA & VS & FL & Acc. & Comp. & Overall \\ \midrule
\multicolumn{1}{c}{(a)} &  &  &  &  & 0.351 & 0.339 & 0.345 \\
\multicolumn{1}{c}{(b)} &  & \ding{52} &  &  & 0.368 & 0.253 & 0.311 \\
\multicolumn{1}{c}{(c)} &  & \ding{52} & \ding{52} &  & 0.363 & \textbf{0.246} & 0.305 \\
\multicolumn{1}{c}{(d)} &  & \ding{52} & \ding{52} & \ding{52} & \textbf{0.335} & 0.258 & \textbf{0.297} \\ \bottomrule 
\end{tabular}
\caption{\textbf{Ablation on DTU evaluation set.} ``LA'' and ``VS'' refer to our late aggregation and view shuffling respectively. ``FL'' is the proposed improved dynamic filtering algorithm~\cite{yan2020dense} scheme. The introduction of each module contributes favorably toward the final results.}
\label{table::abl_in_dtu}
\end{table}
\subsection{Ablation Studies}
In this section, we conduct comprehensive experiments to demonstrate the effectiveness of each contribution.
All experimental methods are trained on the DTU dataset using 5 views and tested on the test set at a resolution of 1152$\times$864.

\noindent{\textbf{The effectiveness of proposed contributions.}} 
We design experiments to analyze the contributions of each module.
As shown in Tab. \ref{table::abl_in_dtu}, the introduction of late aggregation (see (b)) leads to a notable improvement, from 0.345 to 0.311, which is already comparable with some SOTA methods \cite{peng2022rethinking,wang2022mvster}. This demonstrates the effectiveness and significance of preserving view information for depth regression.
The view shuffle strategy (see (c)) based on late aggregation consistently improves the accuracy and completeness of the preserved cost, showing the effectiveness of disentangling view order dependence. The multi-view consistency filter further optimizes the point cloud generation process. The contributions are combined to achieve the best final performance.
\begin{table}[t]
 \centering
 \setlength{\tabcolsep}{7pt}
\begin{tabular}{llll}
\toprule
Methods & Acc.  & Comp. & Overall \\ \midrule
ET~\cite{wang2022mvster}       & 0.380 & \textbf{0.280} & 0.330   \\
CNN~\cite{xu2022learning}     & 0.380 & 0.283 & 0.332   \\
MLP~\cite{sayed2022simplerecon}     & 0.348 & 0.334 & 0.341   \\ \midrule 
\textbf{Ours}    & \textbf{0.321} & 0.303 & \textbf{0.312}   \\ \bottomrule \\
\end{tabular}
\caption{\textbf{Experimented on different aggregation strategies}. 
We compare our late aggregation to different early aggregation methods involving Epi-polar Transformer (``ET'') \cite{wang2022mvster}, CNN-based (``CNN'') \cite{xu2022learning} as well as MLP-based (``MLP'') \cite{sayed2022simplerecon} weight modules. Our method achieves the best performance.
}
\label{tab:aggregation_1}
\end{table}

\noindent{\textbf{Comparison between early and late aggregation.}} 
To perform a fair comparison between early aggregation and late aggregation, we conduct ablation studies under the CasMVSNet~\cite{gu2020cascade} pipeline, using cross-entropy loss function, as well as generic depth filtering and fusion algorithm~\cite{schonberger2016pixelwise}.
For early aggregation, we compare three weight module types: CNN-based~\cite{xu2022learning}, epi-polar transformer based~\cite{wang2022mvster}, and MLP-based~\cite{sayed2022simplerecon} approaches.
As shown in Tab. \ref{tab:aggregation_1}, the late aggregation outperforms the prior early aggregation methods with notable margins, showing its superiority against previous practices.
\begin{table}[t]
 \centering
 \setlength{\tabcolsep}{3.2pt}
\begin{tabular}{lccccc}
\toprule
Methods & $N=3$ & $N=5$ & $N=7$ & $N=9$  \\ \midrule
ET~\cite{wang2022mvster}     &   0.347  & 0.330& 0.332 &  0.335 \\
CNN~\cite{xu2022learning}    &   0.348  & 0.332& 0.336  & 0.333     \\ \midrule
\textbf{Ours}   &\textbf{0.330}  & \textbf{0.312}  &\textbf{0.313}  &\textbf{0.314}           \\
\bottomrule
\end{tabular}
\caption{\textbf{Evaluation on flexible test-time input views.} Our method outperforms the early aggregation methods under different testing views, demonstrating its effectiveness in handling flexible test views.}
\label{tab::viewnumbles}
\end{table}

\noindent{\textbf{Performance on flexible testing image numbers}.
To evaluate the performance with flexible testing views,
we train three models with 5 input views each and test them
with 3, 5, 7, and 9 input views. 
As shown in Tab. \ref{tab::viewnumbles}, we find that all methods show slight performance decreases with 7 and 9 input views. We attribute this to the noise introduced by using more views, as the images in DTU are incorporated in descending quality order.
Despite this phenomenon, our method outperforms the early aggregation methods under different testing views, demonstrating its effectiveness in handling flexible test views.

\begin{table}[t]
 \centering
 \scalebox{0.75}{
 \setlength{\tabcolsep}{1.8pt}
\begin{tabular}{lccccc}
\toprule
Methods      & Acc. & Comp. & Overall & Params(M) & Time(s) \\ \hline
CasMVSNet\cite{gu2020cascade}  &0.325  &0.385   &0.355    &\textbf{0.925}  &  \textbf{0.193} \\
UniMVSNet \cite{peng2022rethinking}  &0.352  &0.278   &0.315    &0.934    &  0.271    \\
TransMVSNet \cite{ding2022transmvsnet} & \textbf{0.312} &0.298   & 0.305   &1.148     &  0.743     \\ 
GeoMVSNet \cite{zhang2023geomvsnet} & \underline{0.331} &\underline{0.259}   & \textbf{0.295}   &15.30     &  0.228     \\ 
\midrule
\textbf{Ours}        & 0.335  &\textbf{0.258}    &\underline{0.297}      &\underline{0.926}  & \underline{0.201}     \\ \bottomrule
\end{tabular}
}
\caption{\textbf{Comparison on network parameters and running time}. Our method achieves comparable performance with the SOTA method \cite{zhang2023geomvsnet}, under fewer network parameters and higher speed.}
\label{table::time_parms}
\end{table}
\noindent\textbf{Comparisons on network complexity and efficiency.}
We compare our method with current state-of-the-art and baseline methods\cite{gu2020cascade,peng2022rethinking,ding2022transmvsnet,zhang2023geomvsnet,cao2022mvsformer} in terms of the number of parameters and running time. 
As shown in Tab. \ref{table::time_parms}. 
Since our method builds upon the plain cascade-based MVS \cite{gu2020cascade} without engineering the network architecture  or pipelines, it yields fewer parameters and less running time.
GeoMVSNet achieves strong performance with relatively heavy parameters (15M) using additional 2D CNN layers.
In contrast, targeting specifically early aggregation issues with memory-efficient operations, our method achieves competitively strong performance with an order of magnitude fewer parameters (0.9M).
\section{Conclusion}
In this paper, we analyze the limitation of a wide-adopted early aggregation scheme and based on our analysis, we introduced a view-preserved cost to achieve late aggregation.
Building upon late aggregation, we have additionally developed a series of techniques to reduce dependence on view order, accommodate versatile test viewpoints, and enhance point cloud filtering.
We aspire that our approach will encourage further research into the design of aggregation strategies for multi-view stereo methods.

\bibliography{aaai24}

\end{document}